\documentclass[10pt,twocolumn,letterpaper]{article}

\usepackage{cvpr}              

\usepackage{algorithm,algorithmicx,algpseudocode}
\usepackage{wrapfig}

\usepackage[dvipsnames]{xcolor}

\definecolor{cvprblue}{rgb}{0.21,0.49,0.74}
\usepackage[pagebackref,breaklinks,colorlinks,citecolor=cvprblue]{hyperref}

\def\paperID{9263} 
\def\confName{CVPR}
\def\confYear{2024}

\title{Freestyle 3D-Aware Portrait Synthesis Based on Compositional \\ Generative Priors}

\author{
Tianxiang Ma\textsuperscript{\rm 1,2}, \ \ 
Kang Zhao\textsuperscript{\rm 3}, \ \ 
Jianxin Sun\textsuperscript{\rm 1,2}, \ \ 
Yingya Zhang\textsuperscript{\rm 3}, \ \
Jing Dong\textsuperscript{\rm 2*} \ \ 
\\
\textsuperscript{\rm 1}School of Artificial Intelligence, UCAS \ \ \ 
\textsuperscript{\rm 2}CRIPAC \& MAIS, CASIA \ \ \ 
\textsuperscript{\rm 3}Alibaba Group \ \ \ 
\\
{\tt\small tianxiang.ma@cripac.ia.ac.cn
}
}

\begin{document}


\twocolumn[{%
\renewcommand\twocolumn[1][]{#1}%
\maketitle
\begin{center}
    \centering
    \captionsetup{type=figure}    \includegraphics[width=1.0\textwidth]{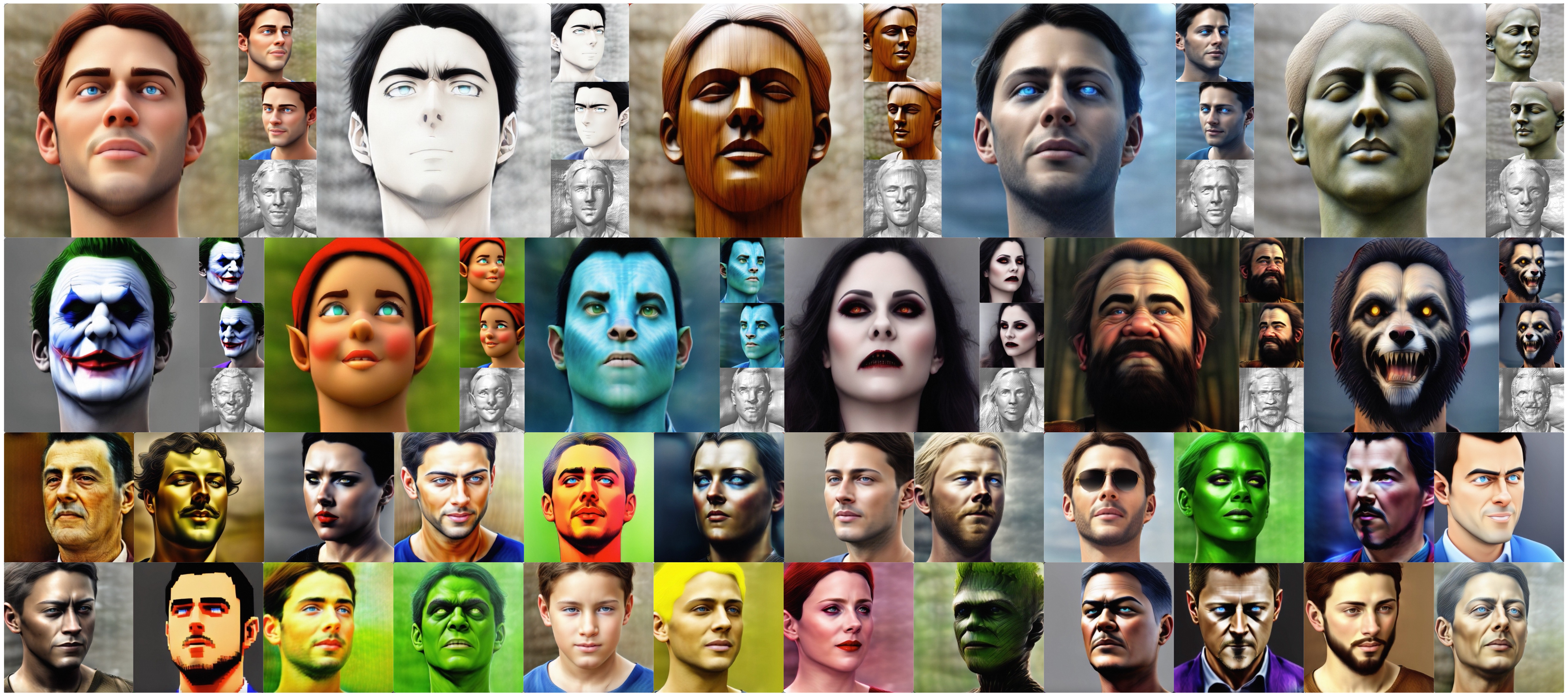}
    \caption{Our freestyle 3D-aware portrait synthesis results. The first row shows man with different styles, the second row shows various characters, and the last two rows show diverse identities with different styles from varied viewpoints. Zoom in for better viewing.}
    \label{teaser}
\end{center}%
}]

\newcommand\blfootnote[1]{%
  \begingroup
  \renewcommand\thefootnote{}\footnote{#1}%
  \addtocounter{footnote}{-1}%
  \endgroup
}
\blfootnote{$^{*}$ Corresponding author.}

\begin{abstract}


Efficiently generating a freestyle 3D portrait with high quality and 3D-consistency is a promising yet challenging task. The portrait styles generated by most existing methods are usually restricted by their 3D generators, which are learned in specific facial datasets, such as FFHQ. To get the diverse 3D portraits, one can build a large-scale multi-style database to retrain a 3D-aware generator, or use a off-the-shelf tool to do the style translation. However, the former is time-consuming due to data collection and training process, the latter may destroy the multi-view consistency. To tackle this problem, we propose a novel text-driven 3D-aware portrait synthesis framework that can generate out-of-distribution portrait styles. Specifically, for a given portrait style prompt, we first composite two generative priors, a 3D-aware GAN generator and a text-guided image editor, to quickly construct a few-shot stylized portrait set. Then we map the special style domain of this set to our proposed 3D latent feature generator and obtain a 3D representation containing the given style information. Finally we use a pre-trained 3D renderer to generate view-consistent stylized portraits from the 3D representation. 
Extensive experimental results show that our method is capable of synthesizing high-quality 3D portraits with specified styles in a few minutes, outperforming the state-of-the-art. 

\end{abstract}

\vspace{-2mm}

\section{Introduction}
Portrait synthesis \citep{karras2019style,karras2020analyzing,gu2021stylenerf,chan2022efficient} is a promising yet challenging research topic for its wide range of application potential, e.g. game character production, Metaverse avatars and digital human. With the rapid development of generative models such as generative adversarial models \citep{goodfellow2020generative}, 2D portrait synthesis has achieved remarkable success. After that, many methods \citep{karras2019style,karras2020analyzing,karras2021alias} are proposed to improve the generation quality to photo-realistic level. 

Recently, 3D portrait synthesis has attracted more and more attention, especially with the emergence of Neural Radiance Field (NeRF) \citep{mildenhall2020nerf}. As the representatives among them, 3D-aware GAN methods \citep{gu2021stylenerf,chan2022efficient,or2022stylesdf} combine NeRF with StyleGANs \citep{karras2020analyzing} to ensure 3D consistency synthesis. 
By mapping an image to the 3D GAN latent space, 3D GAN inversion approaches \citep{ko20233d,lin20223d,yin20223d} can generate or edit a specific 3D portrait. However, both of them fail to create a freestyle 3D portrait, e.g., a style defined by user’s text prompt, since their generators are usually trained on a dataset that follows a particular style distribution, such as the realism style in FFHQ \citep{karras2019style}, which raises a question: how to generate a freestyle 3D portrait at a low cost? One may collect a large number of portrait images with different styles to retrain their 3D generative models, but the data preparing and training process are usually time-consuming. Another potential solution is that synthesizing a dataset-based 3D portrait first, then transferring each portrait view to other styles with a off-the-shelf 2D style transfer tool. Unfortunately, the 3D consistency will be difficult to be maintained.

To this end, we propose an efficient framework to achieve freestyle 3D-aware portrait synthesis in this paper. 
At first, we composite the knowledge of two pre-trained generative priors, EG3D \citep{chan2022efficient} and Instruct-pix2pix (Ip2p) \citep{brooks2022instructpix2pix} we used in this paper, to construct a few-shot stylized portrait dataset with a given style prompt,
avoiding dirty data collection and cleaning. The former generates a multi-view 3D portrait, and the latter performs text-guided style editing in each viewpoint. 
We empirically find the editing results of Ip2p vary significantly along viewpoints for some given style, resulting in the issue of multi-view misalignment.
To alleviate this problem, some optimization strategies are introduced into the inference stage of Ip2p. Secondly, we propose a 3D latent feature generator model, which can map the style information from stylized portrait dataset to 3D latent representation. This generator is first pre-trained with large amount of generative data from EG3D, so that it can learn 3D latent representation well from multi-view portraits. Our 3D latent feature generator enables fast fine-tuning with few-shot stylized portrait dataset, and generates out-of-distribution stylized 3D representation. Finally, we utilize the pre-trained EG3D neural renderer and super-res generator as our 3D Renderer, and we find that this model does not need to be fine-tuned to generate 3D-consistent portrait with a specified style when we have learned a well-stylized 3D implicit representation.
Our high-quality stylized 3D-aware portrait synthesis results are shown in \cref{teaser}, each style is specified by a text prompt, and each 3D portrait model can be fine-tuned in 3 minutes. Our main contributions can be summarized as follows:
\begin{itemize}
    \item We propose a 3D-aware portrait synthesis framework based on compositional generative priors that can be text-driven to generate freestyle 3D avatar.
    \item We further propose a 3D latent feature generator, which allows for quick fine-tuning to map out-of-distribution style to 3D representation. 
    \item Compared with stylized 3D portrait synthesis baselines, our approach has clear advantages in performance and efficiency.
\end{itemize}

\begin{figure*}[t]
\centering
\includegraphics[width=0.8\linewidth]{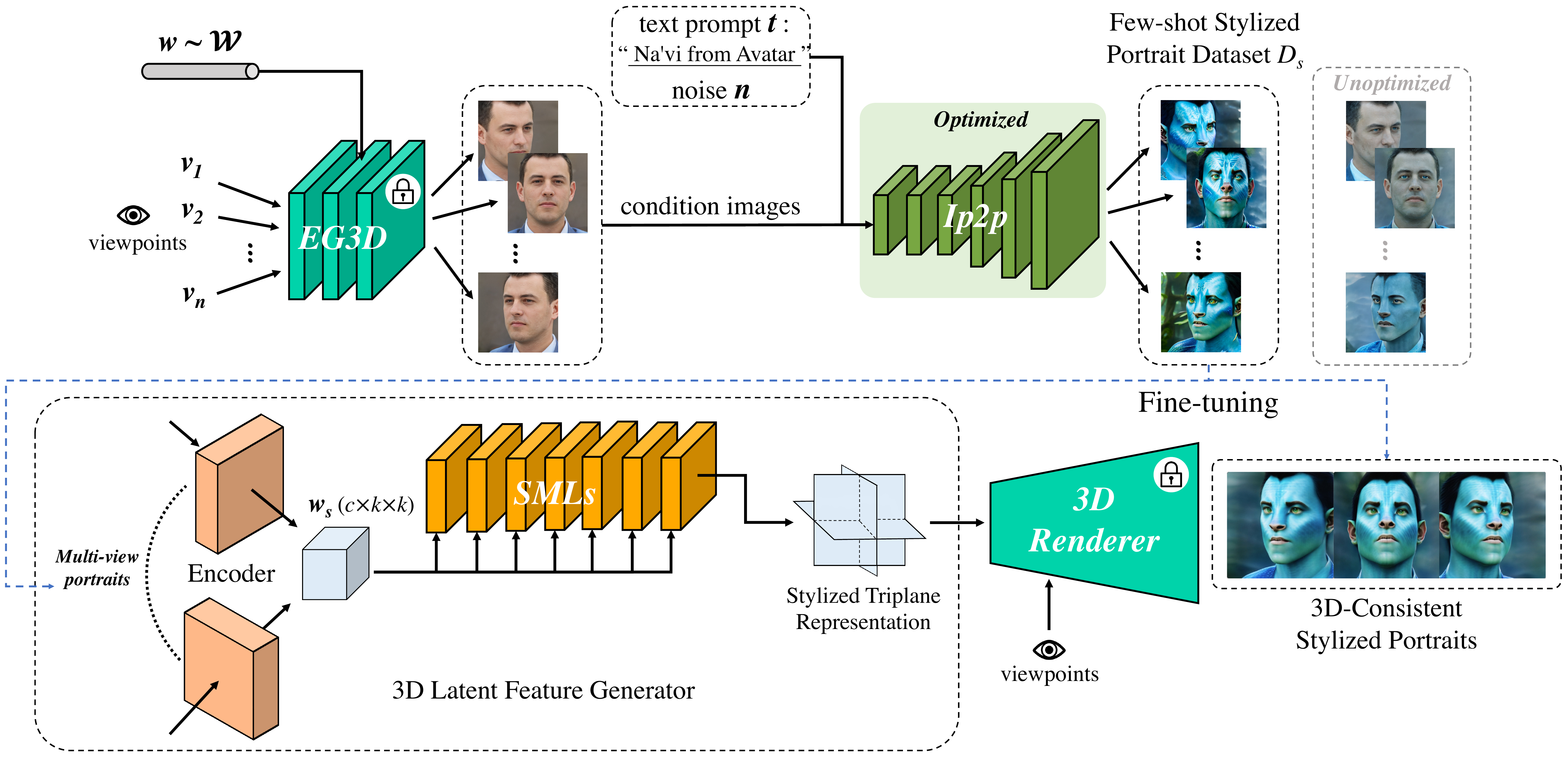}
\caption{The framework of our method. We composite two generative priors EG3D and Ip2p, a 3D-aware GAN generator and a text-guided image editor, to quickly build a few-shot stylized portrait dataset $\mathcal{D}_s$, whose style is specified by the text prompt. Ip2p is optimized to produce more stable and consistent text-guided stylization results. With the constructed few-shot dataset, we quickly fine-tune a 3D latent feature generator to map the style information from the dataset into 3D representation, and then utilize a pre-trained 3D renderer to generate the stylized 3D-consistent portraits.}
\label{framework}
\end{figure*}

\section{Related work}
\subsection{3D-aware GAN}
With the development of neural implicit representation (NIR) represented by neural radiance fields (NeRF) \citep{mildenhall2020nerf}, more and more methods \citep{michalkiewicz2019implicit,niemeyer2019occupancy,chibane2020neural,atzmon2020sal,chabra2020deep,jiang2020local,chibane2020implicit,gropp2020implicit} are focusing on learning 3D scenes and 3D object representation using neural networks. NeRF represents the 3D scene as a series of neural radiance and density fields, and uses volume rendering \cite{kajiya1984ray} technique for 3D reconstruction. Similarly, some methods \cite{sitzmann2019scene,niemeyer2020differentiable} learn neural implicit representation using multi-view 2D images without 3D data supervision. However, even multi-view data is usually expensive to construct in some scenes, such as portraits, so many approaches gradually migrate to learn 3D-aware GAN using unstructured data, i.e., single-view portraits, based on the idea of adversarial training. PiGAN \cite{chan2021pi} proposes a siren-based neural radiance field and uses global latent code to control the generation of shapes and textures. GIRAFFE \cite{niemeyer2021giraffe} proposes a two-stage rendering process, which first generates the low-resolution features with a volume renderer, and then learns to upsample the features with a 2D CNN network. Some methods introduce StyleGAN structures into the 3D-aware GAN. StyleNeRF \cite{gu2021stylenerf} integrates NeRF into a style-based generator to improve rendering efficiency and 3D-consistency of high-resolution image generation. StyleSDF \cite{or2022stylesdf} merges a Signed Distance Fields representation with a style-based 2D generator. EG3D \cite{chan2022efficient} proposes a triplane 3D representation method to improve rendering computational efficiency and generation quality. Some approaches have also started to focus on the control and editing of 3D-aware GANs. FENeRF \citep{sun2022fenerf} and Sem2nerf\citep{chen2022sem2nerf} introduce semantic segmentation into the generative network, and learn a whole neural radiance field with semantic information. CNeRF \citep{ma2023semantic} proposes a compositional neural radiance field to split the portrait into multiple semantic regions, and learns semantic synthesis separately with a local neural radiance field, and finally fuses them into a complete 3D representation of the portrait. Along with the development of 3D-aware GAN, 3D GAN Inversion methods \citep{ko20233d,lin20223d,yin20223d} have appeared. They learn to map real images to the latent space of 3DGAN for image inversion and editing. However, such methods face a problem that they cannot jump out of the pre-trained 3DGAN prior and cannot synthesize out-of-distribution portraits. In this paper, we propose a new framework that can synthesize stylized 3D portraits freely, which is not restricted by the 3D generative prior and can generate 3D portraits of specific styles based on text prompts.

\subsection{Text-guided image editing}
There are numerous image editing methods, and the performance of text-guided image editing methods \citep{avrahami2022blended,hertz2022prompt,kawar2022imagic,brooks2022instructpix2pix,haque2023instruct,liu2023zero} has been qualitatively improved thanks to the advancement of pre-trained image generation large models \citep{rombach2022high,ramesh2021zero,ramesh2022hierarchical} based on the Diffusion model. Ip2p \citep{brooks2022instructpix2pix} is a SOTA text-guided image editing method, which uses two generative priors, GPT-3 \cite{brown2020language} and Stable Diffusion \citep{rombach2022high}, to synthesize a large number of paired images and then train a conditional diffusion model on them. This model allows the users to provide a relatively free text instruction to edit a given image, including stylistic transfer. Therefore, Ip2p is well suitable as a text-guided image editing prior for this paper to perform text-guided style transfer on portraits from different viewpoints. However, the model also has problems, such as poor generation with some simple text prompts and generating portraits with large stylistic variations for different views of the same portrait. We introduce some optimization strategies to alleviate these problems in this paper.

\section{Methodology}

In this section, we detail our freestyle 3D-aware portrait synthesis framework, as shown in \cref{framework}. We first briefly introduce two generative priors, EG3D and Ip2p, and composite them to build a few-shot dataset with a given style. To describe styles more freely, we optimize the inference of Ip2p to make it more stable for stylizing portraits from different perspectives (\cref{subsec:few-shot}). We then use the few-shot dataset to quickly fine-tune our proposed 3D latent feature generator, which is equipped with a pre-trained 3D renderer to generate 3D-consistent stylized portraits (\cref{subsec:I2T}).



\subsection{Few-shot stylized portrait dataset construction} \label{subsec:few-shot}
\noindent \textbf{3D-aware GAN prior.}
As a state-of-the-art 3D-aware GAN method, EG3D \citep{chan2022efficient} can be expressed as $\mathit{G}(\theta, \mathbf{w}, \mathbf{v})$, where $\theta$ is the model parameters, $\mathbf{w}$ is a sampling vector in the $\mathcal{W}$ latent space, and $\mathbf{v}$ is the view direction to be rendered. We randomly sample a $\mathbf{w}$ vector, and set $\mathbf{v}$ as follows: assuming pitch and yaw angles of front view portrait are zero, $\mathbf{v}$ is uniformly sampled $i$ times within both pitch and yaw range of $-30^{\circ}$ degrees to $30^{\circ}$ degrees. We denote the candidate set of $\mathbf{v}$ as $(P, Y)$, which contains $i^2$ sampling results $(i \ pitch \times i \ yaw)$. Then $\mathit{G}$ can output $i^2$ portrait images along each $\mathbf{v}$. Note that these portraits keep the same identity since $\mathbf{w}$ is only sampled once.



\noindent \textbf{Text-guided image editing prior.}
Next, we employ Instruct-pix2pix (Ip2p) \citep{brooks2022instructpix2pix} to perform the style editing on the portraits generated above. Ip2p implements conditional image editing based on Stable Diffusion \citep{rombach2022high}, which can be denoted as $\mathit{T}(\phi, \mathit{I}, \mathbf{n}, \mathbf{t})$, where $\phi$ is the model parameters, $\mathit{I}$ is the input portrait, $\mathbf{n}$ is the Gaussian noise used in the denoising process, and $\mathbf{t}$ is the text prompt that guides the editing direction. Considering different input noise will generate different results, we fix $\mathbf{n}$ to keep the identity unchanged.
Then we let $\mathit{I}$ be the multi-view portraits that are produced by EG3D, and use $\mathit{T}$ to generate stylized portraits. 


\noindent \textbf{Optimizing Instruct-pix2pix inference.}
As mentioned above, some $\mathbf{t}$ will cause Ip2p to generate unsatisfactory stylized portraits.
For example, the style ``Na’vi from Avatar'' will make the portrait style vary greatly from one viewpoint to another. More samples are listed in \cref{subsec:abl}. Therefore, we want to optimize Ip2p to make it generate stable results for different $\mathbf{t}$.
Inspired by SDEdit \citep{meng2021sdedit}, we replace the original Gaussian noise $\mathbf{n}$ with a new noise $\mathbf{n^*}$ during the inference stage of Ip2p:
\begin{equation}
\mathbf{n^*} = Add(\mathcal{E}(I), \mathbf{n}, \tau),
\end{equation}
where $\mathcal{E}(I)$ denotes the latent features obtained from the Stable Diffusion encoder of $I$, $\tau$ is the degree of noise addition, and $Add$ represents the standard DDPM \citep{ho2020denoising} noise addition operation. 
In addition, we design an enhanced prompt to further improve the quality of synthesized portraits: $\mathbf{t}^* = \{\mathbf{t}, \mathbf{t_d}, \mathbf{t_n}\}$, where $\mathbf{t_d}$ and $\mathbf{t_n}$ mean decorative and negative prompts, respectively. Consequently, our stylized portrait generation can be rewritten as: 
\begin{equation}
\mathit{I_s} = \mathit{T}(\phi, \mathit{G}(\theta, \mathbf{w}, \mathbf{v}), \mathbf{n^*}, \mathbf{t}^*),
\end{equation}
where $\mathit{I_s}$ is one stylized portrait. We construct a few-shot stylized portrait dataset $\mathcal{D}_s$ using different view $\mathbf{v}$ from $(P, Y)$, so $\mathcal{D}_s$ contains $i^2$ stylized portraits. The construction pipeline is summarized in Algorithm \ref{algorithm1}.




\subsection{3D Latent Feature Generator} \label{subsec:I2T}
Considering the ability in generating high quality and 3D consistency image, we also utilize EG3D baseline to synthesize stylized 3D portrait. In the experiment, we find that if we have learned a well-stylized 3D implicit representation, the EG3D neural renderer and super-res generator do not need to be fine-tuned to generate 3D-consistent portrait with an out-of-distribution style. Therefore, we utilize the pre-trained EG3D neural renderer and super-res generator as our 3D renderer, and just need to learn a 3D latent feature generator to map the style information from $\mathcal{D}_s$ to the 3D implicit representation. 

There are two other solutions that need to be discussed: 1) Directly fine-tuning the whole EG3D on $\mathcal{D}_s$, which is very inefficient compared to our method and generates somewhat blurred images. 2) Inverting the images from $\mathcal{D}_s$ to the $\mathcal{W}$ latent space of EG3D with GAN inversion methods, which will limit the portrait style within the scope of training set that EG3D is pre-trained, since the triplane feature generator is not changed.


We design the 3D latent feature generator as shown in the \cref{framework}. An Encoder is used to learn the modulation latent feature $\mathbf{w}_s$ from different viewpoints, we expand the latent code $\mathbf{w}_s$ along spatial dimension (from $c$ to $c \times k \times k$) to enrich it with style and structural information. Then we propose the $\textit{SMLs}$ consisting of multiple Stylized modulation layers, which can learn to generate stylized triplane implicit representation. 
For a particular modulation layer in $\textit{SMLs}$, we have the following formula:
\begin{equation}
F_{c, h, w}^{i} = \gamma_{h, w}^{i}(\mathbf{w}_s) \times \frac{F_{c, h, w}^{i}-\mu_{h, w}^{i}}{\sigma_{h, w}^{i}}+\beta_{h, w}^{i}(\mathbf{w}_s),
\end{equation}
where $\mu_{h, w}^{i}$ and $\sigma_{h, w}^{i}$ represent the calculated mean and standard deviation across channel dimension, respectively. $\gamma_{h, w}^{i}$ and $\beta_{h, w}^{i}$ are learnable weight networks. The last layer of $\textit{SMLs}$ outputs the stylized triplane representation.


Although the 3D latent feature generator can be trained on $\mathcal{D}_s$, it still suffers from two challenges:
First, $\mathcal{D}_s$ is a few-shot dataset, it has no more than 100 images for a style in practice.
Second, the portraits in $\mathcal{D}_s$ have more or less differences in details, resulting in 3D inconsistency. 
We solve this problem by pre-training the 3D latent feature generator on the large number of portraits generated by EG3D prior, which can help to build accurate 3D implicit representation from multi-view portraits. 
In particular, in each iteration of the pre-training, we randomly sample to generate an arbitrary viewpoint portrait $\mathit{I_v} \in \mathbb{R}^{512 \times 512 \times 3}$ from EG3D, and record its triplane representation as $\mathbf{p} \in \mathbb{R}^{256 \times 256 \times 96}$, 
Then the pre-training process is constrained by the following loss function:
\begin{equation}
\mathcal{L}_{\textrm{pre} }=\mathbb{E}_{\mathit{I_v}, \mathbf{p}} [ \left\| \mathit{H}(\mathit{I_v}) - \mathbf{p} \right\|_{1} ],
\end{equation}
where $\mathit{H}$ represents our 3D latent feature generator. 




\begin{table}
\vspace{-3mm}
  \begin{minipage}[t]{0.45\textwidth}
  \begin{algorithm}[H]
  \caption{Few-shot dataset construction} \label{algorithm1}
   \begin{algorithmic}[1]
    \State \textbf{Input:} $\mathbf{w} \sim \mathcal{W}, \ \mathbf{n} \sim \mathcal{N}(0,1), \ \mathbf{t},\ \mathcal{D}_s=\emptyset$
    \State $\textbf{for} \ \mathbf{v} \ \textbf{in} \ (P,Y) \ \textbf{do}$
    \State \ \ $I = \mathit{G}(\theta, \mathbf{w}, \mathbf{v})$
    \State \ \ $\mathbf{n^*} = Add(\mathcal{E}(I), \mathbf{n}, \tau)$
    \State \ \ $\mathit{I_s} = \mathit{T}(\phi, I, \mathbf{n^*}, \mathbf{t}^*)$
    \State \ \ $\mathcal{D}_s = \mathcal{D}_s \cup \mathit{I_s}$
    \State \textbf{end for}
     \end{algorithmic}
     \end{algorithm}
  \end{minipage}%
  \hfill
 \begin{minipage}[t]{0.45\textwidth}
    \begin{algorithm}[H]
    \caption{The 3D latent feature generator fine-tuning} \label{algorithm2}
    \begin{algorithmic}[1]
    \State \textbf{Input:} $\mathcal{D}_s$, Pre-trained 3D latent feature generator
    \State Init 3D generator net using the pre-trained model
    \State \textbf{repeat}
    \State \ \ Randomly select $\mathit{I^1_v}, \mathit{I^2_v}$ from $\mathcal{D}_s$
    \State \ \ Fine-tuning 3D latent feature generator using loss: 
    \State \ \ \ \ \ \ $\mathcal{L}_{\textrm{total} } = \lambda_{\textrm{rec}} \mathcal{L}_{\textrm{rec}} + \lambda_{\textrm{dr}} \mathcal{L}_{\textrm{dr}} $ 
    \State \textbf{until} end of iterations 
  \end{algorithmic}
\end{algorithm}
  \end{minipage}
\end{table}

\begin{figure*}[t]
\centering
\includegraphics[width=0.89\linewidth]{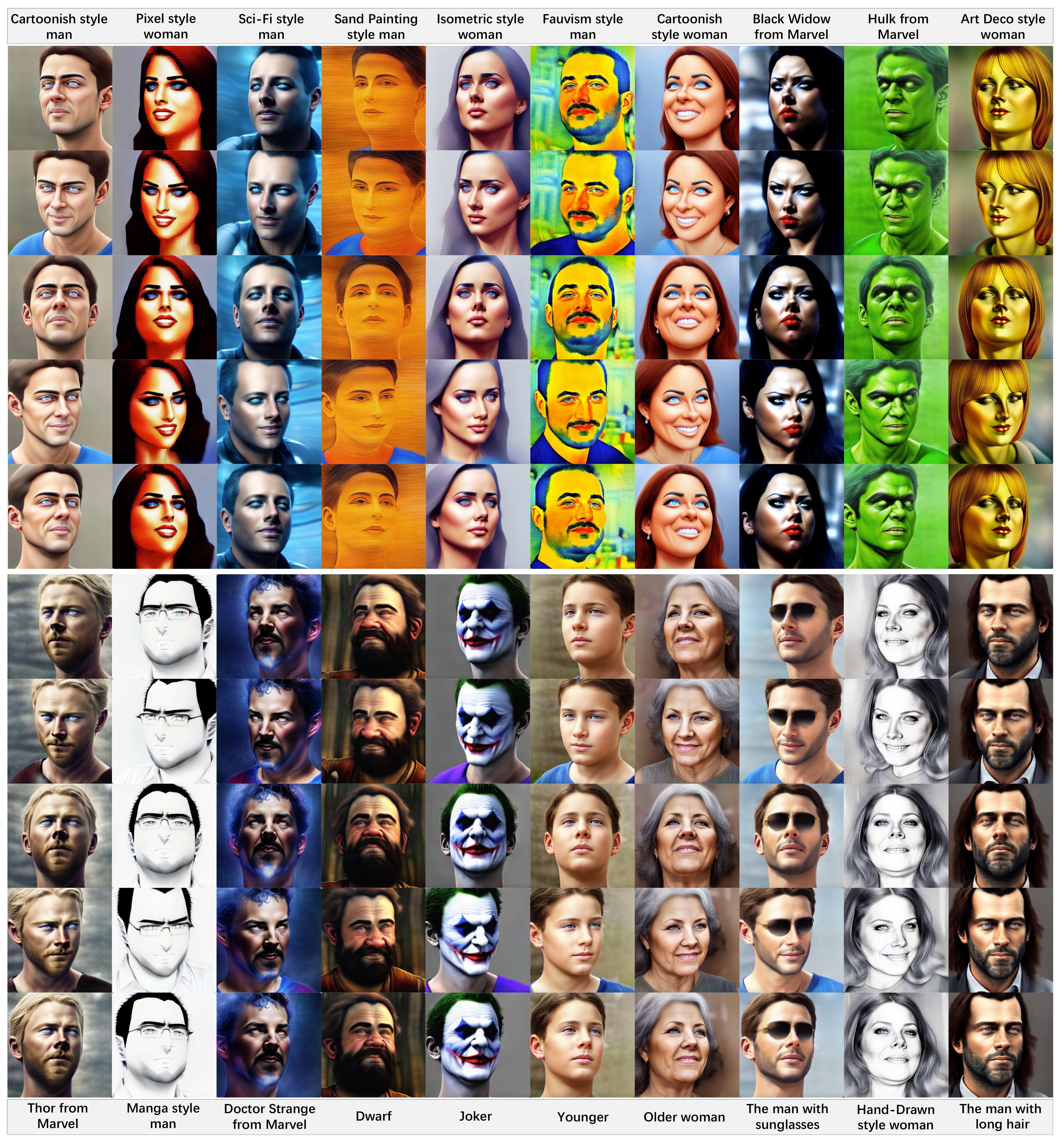}
\caption{Multi-style and Multi-identity 3D portrait synthesis results. Zoom in for better viewing.}
\label{stylized_portraits}
\end{figure*}

\begin{figure*}[t]
\centering
\includegraphics[width=0.89\linewidth]{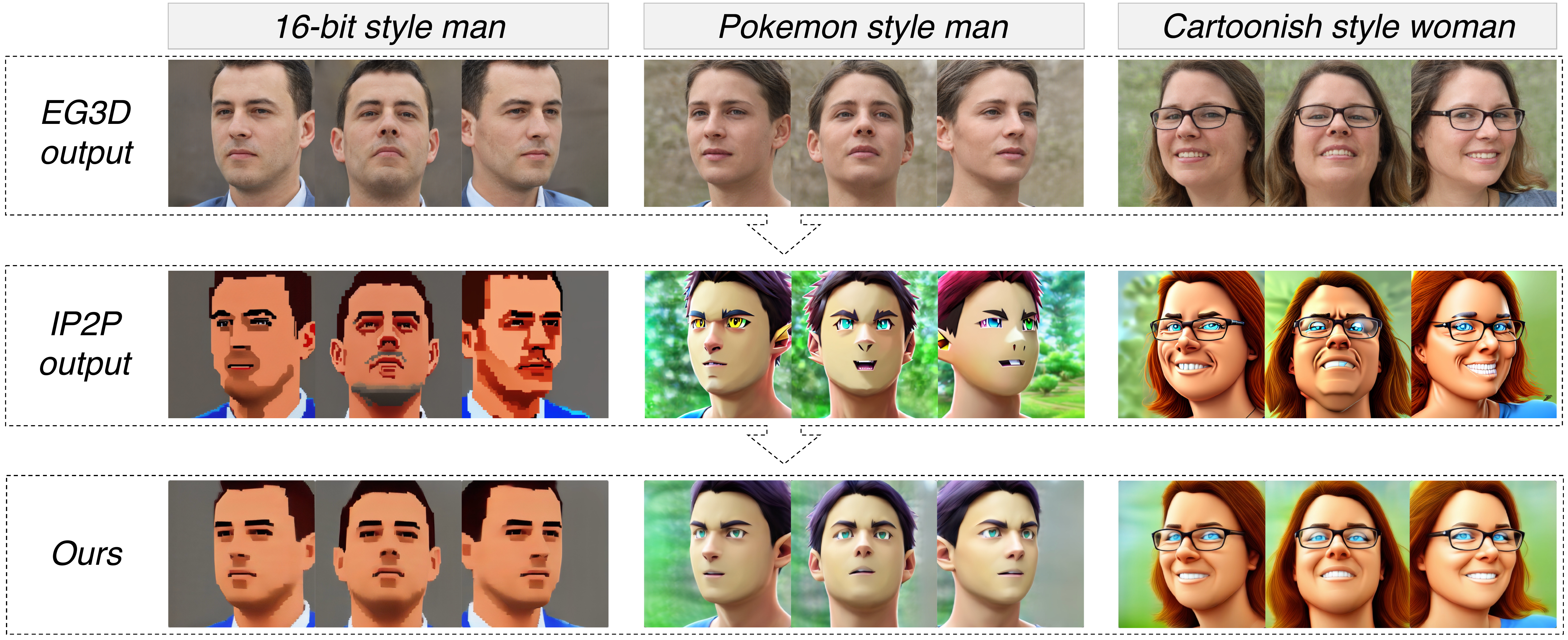}
\caption{The outputs of different stages in our method.}
\label{three_stage}
\end{figure*}

\begin{figure*}[t]
\centering
\includegraphics[width=0.89\linewidth]{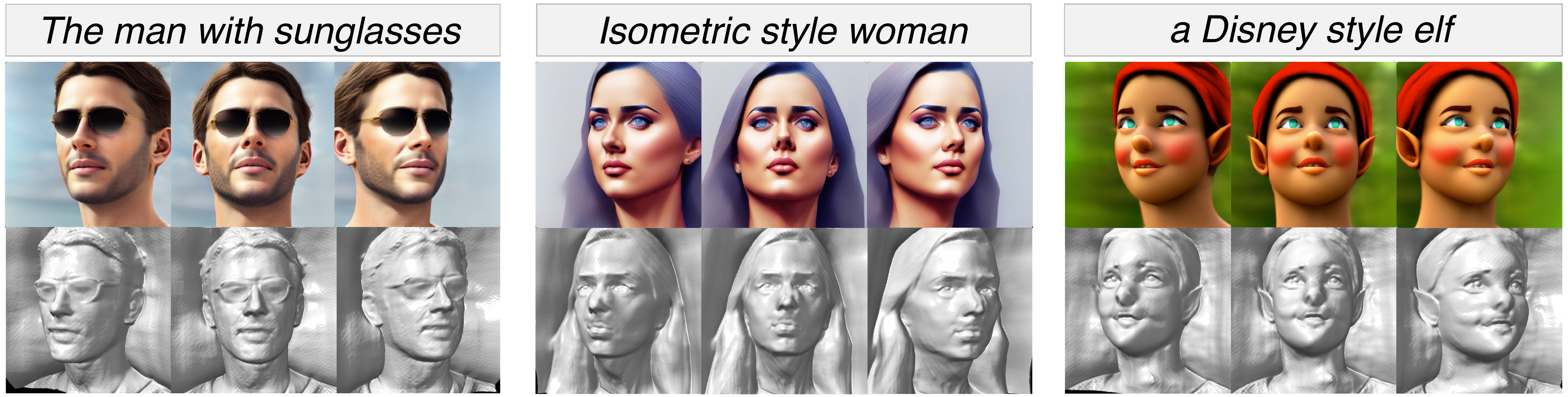}
\caption{Visualization of 3D geometry from different stylized generation results.}
\label{shape}
\end{figure*}

\begin{figure*}[ht]
\centering
\includegraphics[width=1.0\linewidth]{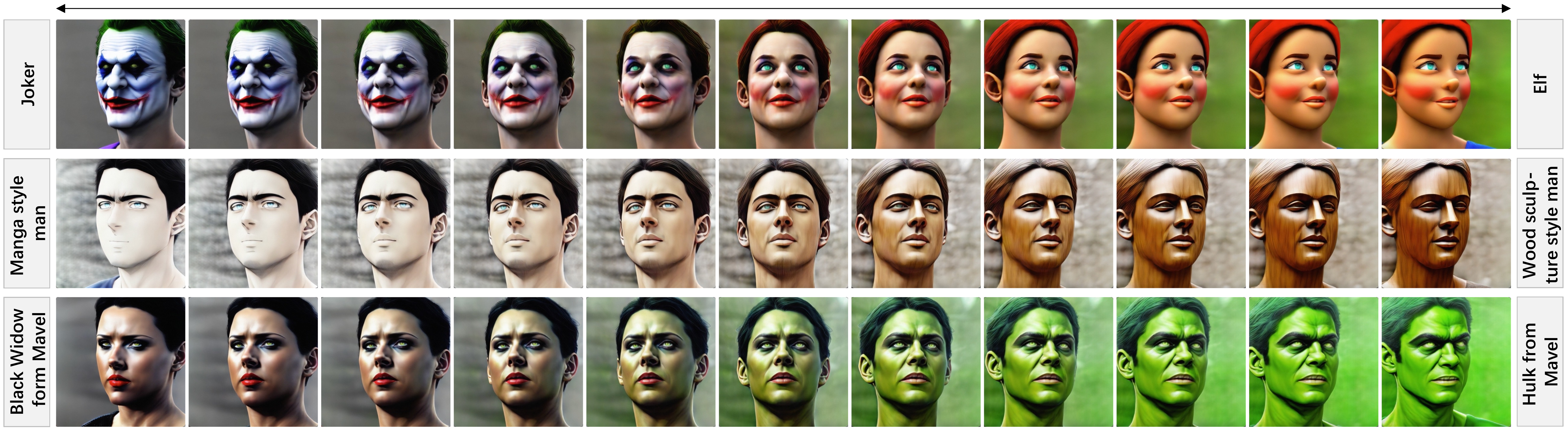}
\caption{The 3D portrait results synthesized by mixing different styles in the $\mathbf{w}_s$ latent space.}
\label{stylemix}
\end{figure*}

\noindent \textbf{Training.} 
After being pre-trained, the 3D latent feature generator can be quickly fine-tuned on the few-shot stylized portrait dataset $\mathcal{D}_s$.
We use a multi-view cross-construction strategy for learning a more robust 3D implicit representation.
For two arbitrary viewpoint portraits $\mathit{I^{1}_v}$ and $\mathit{I^2_v}$ in $\mathcal{D}_s$, we use the following reconstruction loss:
\begin{equation}
\begin{aligned}
\mathcal{L}_{\textrm{rec}} = \mathbb{E}_{\mathit{I^1_v},\mathit{I^2_v} \in \mathcal{D}_s} &[ \left\| \mathit{G^*}(\mathit{I^1_v}, \mathbf{v_2}) - \mathit{I^2_v} \right\|_{1} \\
& + lpips(\mathit{G^*}(\mathit{I^1_v}, \mathbf{v_2}), \mathit{I^2_v}) ],
\end{aligned}
\end{equation}
where $\left\| \cdot \right\|_{1}$ is the $L_1$ reconstruction loss, $\mathit{G^*}$ represents the whole model with 3D latent feature generator and 3D renderer, and the parameters are input image and the rendering viewpoint, respectively. $lpips(\cdot,\cdot)$ is the Learned Perceptual Image Patch Similarity \citep{zhang2018perceptual} loss, which calculates the distance of the latent features extracted from the VGG network. In addition to the reconstruction loss, we add the density regularization, which encourages smoothness
of the density field rendered by triplane and prevents sharp or hollow portrait shapes during fine-tuning. The density regularization loss is shown as follows,
\begin{equation}
\mathcal{L}_{\textrm{dr}}=\mathbb{E}_{\mathbf{x},\delta} [ \left\| \sigma(\mathbf{x}) - \sigma(\mathbf{x} + \delta \cdot \mathbf{x}) \right\|_{1} ],
\end{equation}
where $\mathbf{x}$ are the random sampling points in the volume rendering, $\delta$ is a small Gaussian noise, and $\sigma(\mathbf{x})$ denotes the density rendering process. Thus the final loss function is: 
\begin{equation}
\mathcal{L}_{\textrm{total} } = \lambda_{\textrm{rec}} \mathcal{L}_{\textrm{rec}} + \lambda_{\textrm{dr}} \mathcal{L}_{\textrm{dr}},
\end{equation}
where $\lambda_{\textrm{rec}}$ and $\lambda_{\textrm{dr}}$ are loss weights. The fine-tuning process is listed in Algorithm \ref{algorithm2}.

\section{Experiments}

\subsection{Implementation detail}
Our method is implemented in PyTorch using an NVIDIA A100. We use Adam optimizer with learning rate of $0.002$ and $\beta_1 = 0, \beta_2 = 0.99$. The number of samples $i$ in few-shot dataset construction is 10. Other parameters, such as camera focal length, use the EG3D default settings. In Ip2p inference, we set the number of time step $\mathbf{s}$ in DDIM \citep{song2020denoising} to 20. The degree of noise addition $\tau$ is 0.9, and the image guide and text guide weight parameters of Ip2p are set to 1.5 and 20.0. The decorative prompt $\mathbf{t_d}$ is ``realistic, detail, 8k, photorealistic", then the positive prompt input for Ip2p is ``turn the head into $\mathbf{t}$, $\mathbf{t_d}$". The negative prompt $\mathbf{t_n}$ is ``unclear facial features, non-face objects, ugly, bad". Note that we fix $\mathbf{t_d}$ and $\mathbf{t_n}$, and only change $\mathbf{t}$ in all following experiments. Fine-tuning $\mathbf{t_d}$ or $\mathbf{t_n}$ will polish the generated results, but it's not the scope of this paper. For each text prompt $\mathbf{t}$ we randomly sample a $\mathbf{w}$ and construct the few-shot dataset according to Algorithm \ref{algorithm1}. In 3D latent feature generator, The spatial dimension $k$ of $\mathbf{w}_s$ is 32. $\textit{SMLs}$ consists of 7 modulation layers. The 3D generator is pre-trained on EG3D randomly sampled data for 100k iterations. When fine-tuning the model on the few-shot dataset $\mathcal{D}_s$, the loss weights are set as $\lambda_{\textrm{rec}}=10.0$ and $\lambda_{\textrm{dr}}=0.2$.

\subsection{freestyle 3D portrait synthesis}
In this section, we show the 3D portrait synthesis results of our approach. As shown in \cref{stylized_portraits}, 
our method can generate diverse style 3D portrait, and the synthesized portraits are high quality and 3D consistent, proving our ability of freestyle generation. 
Then, 
we display the outputs of different stage (i.e., EG3D, IP2P and final output) in
\cref{three_stage}, 
it can be seen EG3D outputs the original 3D portrait, IP2P will add the style into the portrait images (the facial details and consistency can not be guaranteed), our method will generate the final high quality and 3D-consistency results.
In addition, we generate the 3D geometry results of different style in \cref{shape}, which can accurately reflect the given style. 
Last but not least, \cref{stylemix} shows the smooth style mixing results of our method. We interpolate the encoded features of two different style in the $\mathbf{w}_s$ latent space, and use the new triplane implicit representation to synthesize 3D portraits. The results reveal the controllability and potential style editing ability of our model.



\subsection{Comparison experiments}

\begin{figure}[t]
\centering
\includegraphics[width=1.0\linewidth]{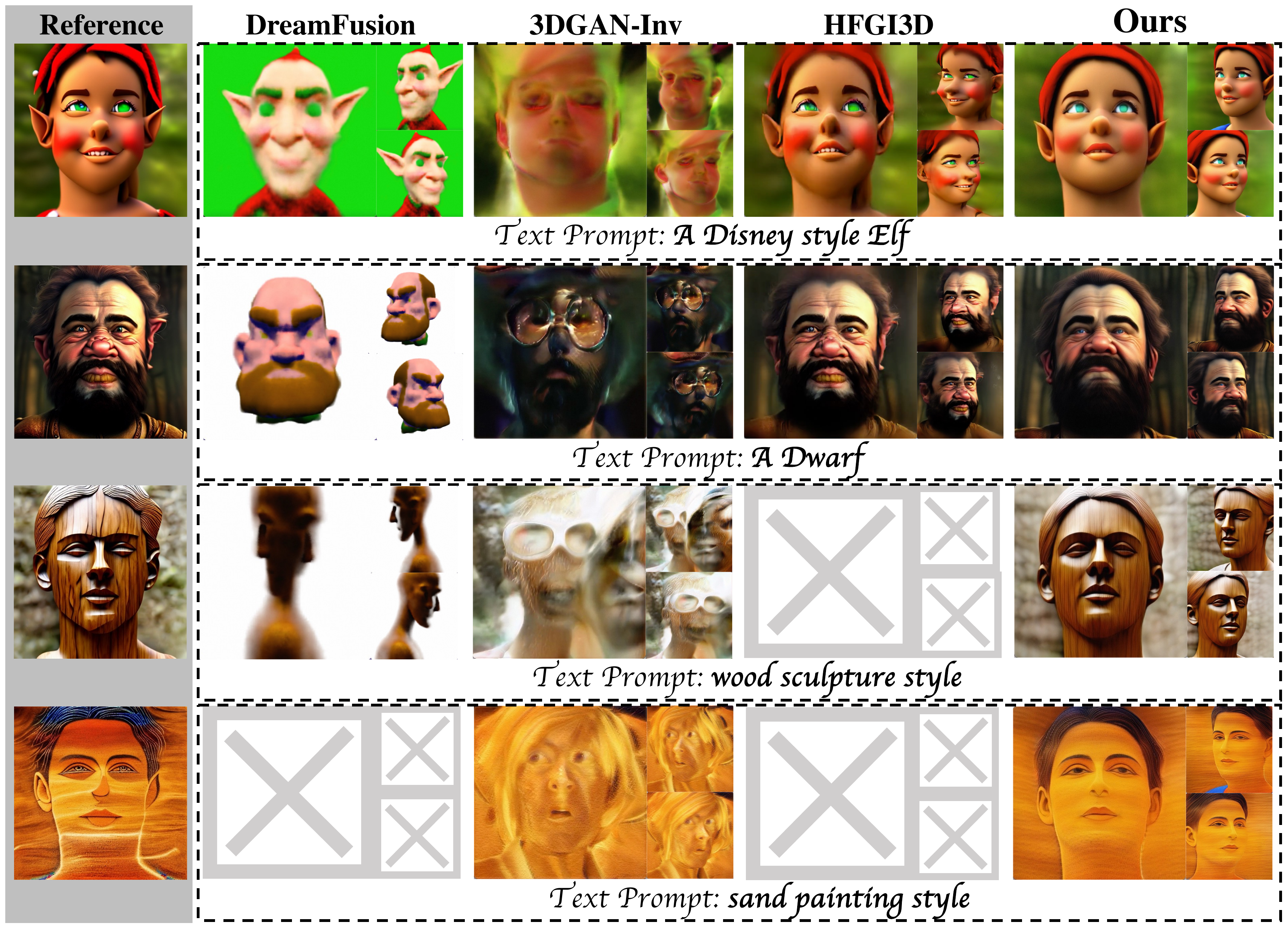}
\caption{Qualitative comparison results between our method and baselines. Reference image represents the result after Ip2p style transformation, which is used as the input of the 3D GAN Inversion methods. The part with no result means that the method is unable to generate a image.}
\label{compare}
\end{figure}


\begin{table*}[t]
  \caption{User Study, quantitative evaluation and running time of different methods.}
  \centering
  \begin{tabular}{cccc|c|c}
    \toprule
    method & Text-image Similarity & Image Quality & 3D Consistency & CLIP Score & Average Time\\
    \midrule
    DreamFusion \citep{poole2022dreamfusion} & 2.82 & 2.24 & 3.18 & 0.304 & $\sim$ 40 mins\\
    3DGAN-Inv \citep{ko20233d} & 2.04 & 1.90 & 2.59 & 0.229 & $\sim$ 4 mins \\
    HFGI3D \citep{Xie2023high} & 3.40 & 3.54 & 3.67 & 0.310 & $\sim$ 10 mins \\
    Ours & \textbf{4.05} & \textbf{4.27} & \textbf{4.34} & \textbf{0.332} & $\sim$ 3 mins \\
    \bottomrule
  \end{tabular}
  \label{quantitative_evaluation}
\end{table*}

\noindent \textbf{Baselines.} We divide the baselines into two categories. 1) Text-to-3D. DreamFusion\footnote{\url{https://github.com/ashawkey/stable-dreamfusion}} \citep{poole2022dreamfusion} is a representative method that generates 3D images based on the text prompts. 2) Image style transfer + 3D GAN Inversion. 3DGAN-Inv \citep{ko20233d} and HFGI3D \citep{Xie2023high} are two SOTA methods of 3D GAN inversion. We use them with Instruct-pix2pix as the baselines.


The results of our method compared with baselines are shown in \cref{compare}. DreamFusion, the representative of the Text-to-3D method, is able to optimize the model to synthesize 3D portraits based on text prompts, such as "A Disney style Elf", but the generated results have low quality, while for more specific portrait styles, such as "sand painting style", no results can be synthesized. What's more, the portraits with large stylistic variations cannot be inverted well using the 3DGAN-Inv, because the 3D portraits synthesized by this method cannot escape from the domain of the pre-trained 3D-aware GAN model. The results of HFGI3D are better, but the 3D shape is destroyed after inversion, as shown in the Elf's face. At the same time, for some more difficult samples, this method cannot produce effective results, such as the last two lines.
In contrast, our method is able to synthesize high quality 3D portraits that satisfy both 3D consistency and stylization. 

For quantitative evaluation, we conduct a user study. We invite 30 volunteers to evaluate each method from three perspectives, namely the text-image similarity, the quality of the generated images, and the 3D consistency of the results. Each item is scored on a 5-point scale, and the average is calculated as the final result. As shown in \cref{quantitative_evaluation}, our approach achieves the best scores under each dimension. We also calculate the CLIP Score for each method shown in \cref{quantitative_evaluation}, which uses the CLIP \citep{radford2021learning} model to extract features and calculate cosine similarity between input text and generated image. Our method also achieves optimal result.
In addition, our method also has a significant advantage in running time, as shown in \cref{quantitative_evaluation}. Our method is able to fine-tune model on the given text prompt in about 3 minutes, while other approaches require more time. Although 3DGAN-Inv costs comparable running time to ours, its portrait generation quality is poor and fails to generate freestyle portraits.



\begin{figure}
\centering
\includegraphics[width=1.0\linewidth]{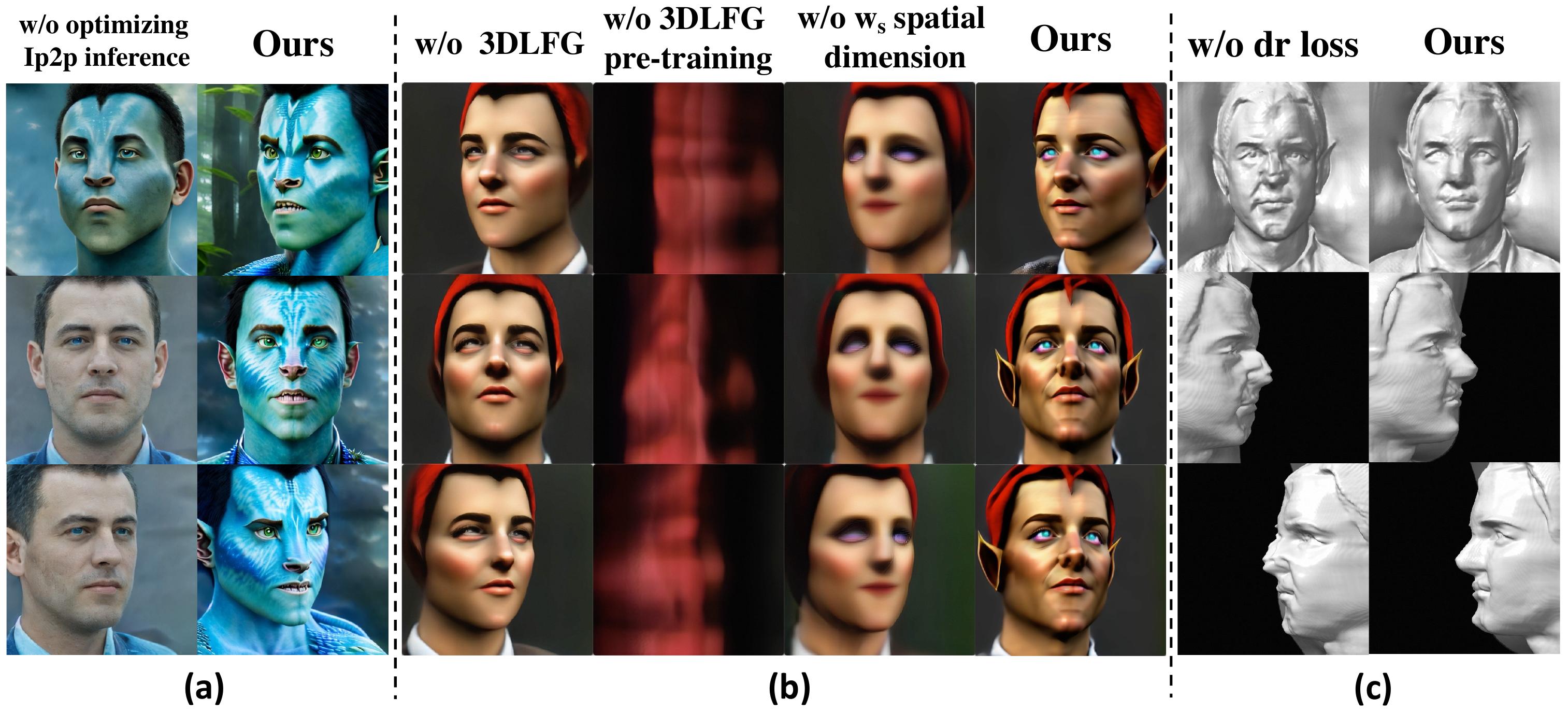}
\caption{The ablation studies of our approach. (a) Ip2p generated results w/ or w/o inference optimization. (b) Different ablation experiments on the 3D Latent Feature Generator (3DLFG). (c) 3D shape outputs w/ or w/o dr loss.}
\label{ablation}
\end{figure}

\subsection{Ablation study} \label{subsec:abl}


\begin{table}[t]
  \caption{Quantitative ablation results on 3D latent feature generator.}
  \centering
  \begin{tabular}{cc}
    \toprule
    method & CLIP Score \\
    \midrule
    w/o 3D generator & 0.251  \\
    w/o pre-training  & 0.113 \\
    w/o $\mathbf{w}_s$ spatial dimension  & 0.282  \\
    Ours & \textbf{0.332} \\
    \bottomrule
  \end{tabular}
  \label{clip_score}
\end{table}


We conduct the ablation studies of our method from three aspects, which are shown in \cref{ablation}. First, when we do not optimize the Ip2p inference, some character prompts generate poor results, such as the example in (a) of \cref{ablation}). Second, we perform the ablation of the 3D latent feature generator in (b) of \cref{ablation}. 1) When we remove the 3D latent feature generator and directly fine-tuning the whole EG3D, 
we can achieve some stylization effect. However, the generation quality is somewhat poor, and this fine-tuning is inefficient, taking dozens of minutes. 2) When using our 3D latent feature generator without pre-training, training from scratch on the few-shot dataset will not be able to learn the correct 3D implicit representation. Because it is difficult to establish the mapping from the image to the triplane representation only using 3D inconsistent data. 3) When the $\mathbf{w}_s$ code of our 3D latent feature generator is a vector without spatial dimension, the learned 3D portrait is blurred and lacks details. Furthermore, we calculate the CLIP Score of different ablation models in \cref{clip_score}, and our method achieves the best result. 
In addition, we ablate the density regularization (dr) loss used in model training, shown in (c) of \cref{ablation}. When the density regularization is not used, the synthesized portrait shape will have rough surfaces.

\section{Limitation}
\begin{wrapfigure}{r}{0.5\linewidth}
    \includegraphics[width=\linewidth]{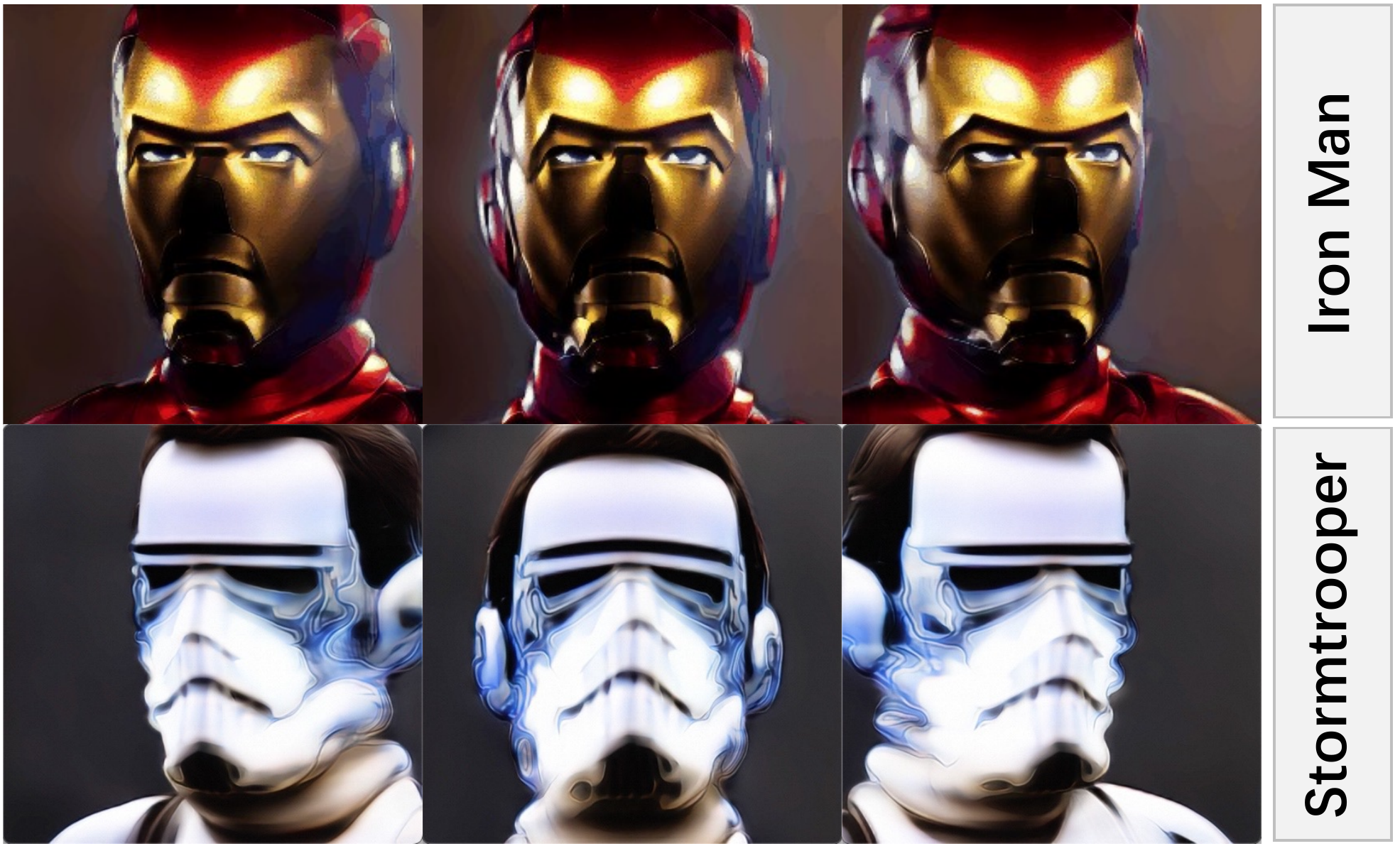}
    \vspace{-5mm}
    \caption{\small Some bad cases.}
    \label{hard_case}
\end{wrapfigure}
Our approach is based on two powerful pre-trained generative priors, which are the basis of our method's ability to synthesize high-quality stylized 3D portraits. At the same time, our method is limited by both priors, especially Ip2p, which is unable to achieve perfect 3D-consistent portrait stylization in different viewpoints, so the final synthesized 3D portrait of our method differs slightly from the style generated by Ip2p. Meanwhile, some stylistic changes that differ significantly from the human portrait shape, such as "Iron Man" and "Stormtrooper" in \cref{hard_case}, result in poorer quality 3D portraits compared to human portraits. Because when Ip2p generates this type of style, the stylization effect will be more different under different views.

\section{Conclusions}
This paper proposes a novel freestyle 3D-aware portrait synthesis framework based on compositional generative priors. We combine two pre-trained generative priors, EG3D and Ip2p, to quickly construct the out-of-distribution stylized multi-view portraits. We optimize Ip2p inference in order to stylize portraits at different views more freely and stable.
We propose a 3D latent feature generator, which learns to map the style information from few-shot stylized portrait dataset to the 3D implicit representation. Equipping the generator with a pre-trained 3D renderer, we can generate 3D-consistent stylized portraits.
A large number of high-quality 3D portrait synthesis results and comparison experiments with baselines show the superiority of our method. 


{
    \small
    \bibliographystyle{ieeenat_fullname}
    \bibliography{main}
}


\end{document}